# Cancer Cell Classification using Deep Learning

Dr.Chayadevi M L
Computer Science Engineering
*BNM Institute of Technology*
*Bangalore, India*
hodcse@bnmit.in

Nidhi Srivastava
Computer Science Engineering
*BNM Institute of Technology*
*Bangalore, India*
srivastavanidhi0212@gmail.com

Rakshith Mahishi
Computer Science Engineering
*BNM Institute of Technology*
*Bangalore, India*
rakshith911@gmail.com

Praneeth Kumar T
Computer Science Engineering
*BNM Institute of Technology*
*Bangalore, India*
praneeth.kumt@gmail.com

*Abstract—* In the current technological era, the medical profession has emerged as one of the researchers' favorite subject areas, and cancer is one of them. Because there is now no effective treatment for this illness, it is a matter of concern. Only if this disease is discovered early may patients be rescued (stage I and stage II). The likelihood of survival is quite low if it is discovered in later stages (stages III and IV). The application of machine learning, deep learning, and data mining techniques in the medical industry has the potential to address current issues and bring benefits. Numerous symptoms of cancer exist, including tumors, unusual bleeding, increased weight loss, etc. It is not necessary for all tumor types to be cancerous. There are two sorts of tumors: benign and malignant. To give patients, the right care, symptoms must be carefully examined, and an automated system is to distinguish between benign and malignant tumors. Most data produced in today's online environment comes from websites related to healthcare or social media. Using data mining techniques, it is possible to extract symptoms from this vast amount of data, which will be helpful for identifying or classifying cancer. This research classifies bacteria cells as benign or cancerous using various Deep Learning Algorithms. To get the best and most reliable results for the classification, a variety of methodologies and models are trained and improved.

## I. INTRODUCTION

A collection of disorders is called cancer. It entails unnatural cell development. It spreads quickly and has an impact on other bodily parts. Not every type of tumor is necessarily malignant. Some tumors do not spread throughout the body. Numerous symptoms of cancer include tumors, unusual bleeding, persistent coughing, increased weight loss, etc. Nearly 100 different varieties of cancer can afflict a person. Cancer research is one of the difficult, alluring, and important areas of attention in the medical industry. Accurate automatic tumor and cancer prediction systems are required in order to give patients the proper care. The earlier therapies were manual and clinically based.

The following are some drawbacks of these traditional categorization methods:

- Slow diagnosis procedure

- Tumor/cancer prognosis based on pathology reports, which may raise the patient's efficacy

- Different kinds of clinical courses are needed for cancer classification and prediction.

Various machine learning methods are utilized in the contemporary world to address various issues and design automated systems. These technologies are able to identify and categorize cancer in photos or patient information like age and symptoms. The field of cancer research is not new to the use of machine learning.

Since the last 20 years, decision trees and artificial neural networks (ANNs) have been widely used in various industries for early cancer detection and diagnosis. One of the recent PubMed surveys offers data on the amount of research done on cancer detection techniques. It demonstrates that papers based on the idea of the relationship between cancer and machine learning are published in numbers of at least 1500. Many studies focus on using machine learning techniques for identifying, classifying, and detecting cancer or tumors. Machine learning was once exclusively beneficial for detecting and diagnosing cancer, but more recent algorithms have mostly concentrated on cancer prognosis and prediction. The most popular machine learning techniques are data mining algorithms, which are used to classify gene expression data and cancer.

The purpose of the scenario and cancer prophecy is to:

1) The first step in predicting cancer susceptibility is to determine the likelihood that a certain type of cancer will develop before it manifests itself.

2) After giving a remedy for the original cancer, the extrapolation of cancer recurrence determines the likelihood of cancer development again.

3) After a cancer diagnosis, the likelihood of cancer survival is determined by looking at factors including life expectancy, survival, progression, tumour drug sensitivity, etc.

In the modern internet age, websites for healthcare or social media create a lot of data. Data mining techniques can be used to extract symptoms from this vast amount of data, which will be further helpful for cancer detection or classification. The two most crucial methods utilised for symptom analysis and cancer classification are clustering and classification. This essay provides an overview of recent studies that use both online and offline data to classify cancers. This survey covers the most recent findings in the cancer prognosis and prediction.

In this paper, Section II will discuss the literature review, Section III will describe the proposed system, and Section IV will provide a conclusion.

## II. LITERATURE SURVEY

Epimack Michael et al. proposed A framework for classifying breast cancer using machine learning that has been optimized for improved performance. This paper takes a different stance in setting the values of the hyperparameters. They propose an automatic ML that can automatically build optimized ML algorithms rather than using default values.

This dataset consists of ultrasound images featuring 600 benign lesions and 312 malignant lesions. To further, an outlier detection algorithm and segmentation based on binarization using openCV to obtain the region of interest. Noise is removed using wavelet filter, an inbuilt filter of PyRadiomics toolbox implemented in python. 185 features were extracted out of which 15 of them were used. Finally, these processed images were passed through the five ML classifiers, namely, support vector machine (SVM), k-nearest neighbor (k-NN), random forest (RF), XGBoost, and LightGBM. Overall, the LightGBM classifier achieved the highest performance in terms of accuracy (99.86%), precision (100%), recall (99.60%), and F1 score (99.80%) compared to the other classifiers.

Epimack Michael et al. [1] proposed a machine learning framework for breast cancer classification that is optimized to improve performance. The proposed framework involves an automatic method for setting hyperparameter values, which allows for the creation of optimized machine learning algorithms. The study uses a dataset of 912 breast ultrasound images, including 600 benign and 312 malignant lesions. Outlier detection and image segmentation techniques are applied to the data to identify the region of interest, and noise is removed using a wavelet filter. A total of 185 features are extracted, and 15 of them are selected for use in the analysis. The processed images are then input into five different machine learning classifiers: support vector machine (SVM), k-nearest neighbor (k-NN), random forest (RF), XGBoost, and LightGBM. The results show that the LightGBM classifier performs best, achieving an accuracy of 99.86%, precision of 100%, recall of 99.60%, and an F1 score of 99.80%.

Zeebaree et al. [2] developed a computer-aided diagnosis (CAD) system for breast cancer using machine learning and region-growing segmentation based on morphological characteristics in breast ultrasound images. A key aspect of this work is the creation of an automatic, trainable model that is able to accurately segment and classify breast cancer in ultrasound images. To capture different structures, intensities, and locations, the model uses multiple texture features including 7 moments, FD, and HOG rather than just one feature. The model was trained on a dataset of 250 ultrasound images, including 100 benign and 150 malignant lesions, and achieved an accuracy of 93.1% for detecting malignant lesions and 90.4% for benign lesions when using an artificial neural network (ANN) for classification. To eliminate the under-segmentation issue, region growing strategy us used.

Majid Nawaz et al. [3] proposed a deep learning-based CNN model for multiclass breast cancer classification approach that aims to not only distinguish between benign and malignant tumors, but also predict the subclass of the tumor, such as Fibroadenoma or Lobular carcinoma. The proposed approach was tested on histopathological images from the Break His dataset and found to achieve high performance, with 95.4% accuracy in the multi-class breast cancer classification task using a DenseNet convolutional neural network (CNN) model. This performance is competitive with state-of-the-art models. This paper utilized modified Denset model and applied transfer learning methodology. Since the denset is more suitable for non-microscopic images, kernel of 7x7 sizes for the first convolutional layer was used to detect small variation and substance in the image and extract more important features.

This study [4] develops a new deep learning (DL) model based on the transfer learning (TL) technique to effectively support the automatic detection and diagnosis of the BC suspicious region based on two techniques, namely 80-20 and cross-validation. DL architectures are designed to focus on particular problems. TL applies the knowledge acquired while resolving one issue to another pertinent issue. The proposed model for mammographic image analysis uses pre-trained convolutional neural network (CNN) architectures, including Inception V3, ResNet50, Visual Geometry Group networks (VGG)-19, VGG-16, and Inception-V2 ResNet, to extract features from the MIAS dataset. By classifying mammogram breast images Experimental results show the effectiveness of the proposed model, with overall accuracy, sensitivity, specificity, precision, F-score, and AUC of 98.96%, 97.83%, 99.13%, 97.35%, 97.66%, and 0.995, respectively, when using the 80-20 method, and 98.87%, 97.27%, 98.2%, 98.84%, 98.04%, and 0.993 when using the 10-fold cross-validation method.

A computer-aided diagnostic (CAD) [5] approach for classifying patients into three categories (cancer, no cancer, and non-cancerous) under the control of a database has been proposed in this study. The classification stage has been analysed using Convolution Networks, Support Vector Machines (SVM), and Random Forest, three efficient classifiers (RF). The segmented thermographic pictures have been subjected to analysis and classification using a Convolutional Neural Network (CNN). This method divides the image input variables into pre-expanded RGB and Gray channels, which are then combined with an autonomous image denoising and classification. The resulting output will feed both processes to the breast image analysis and feature extraction network, where a two-extraction determines whether the image is benign or malignant. The accuracy that CNN attained was 99.67 percent, compared to the accuracy that SVM and RF attained, which were both 89.84 percent and 90.55 percent, respectively.

This paper [6], presents a deep learning-based technique for the segmentation and detection of colorectal cancer from digitized H&E-stained histology slides. When compared to pathologist-based diagnosis using the same type of slides, the neural network approach achieved a median accuracy of 99.9% for normal slides and 94.8% for cancer slides. Digital slides pertaining to 307 cases of colorectal cancer have been acquired. 222 slides had varied percentages of colorectal cancer, while 85 slides came from normal colon tissue. The entire slide is broken up into several patches, each of which is called separately. Assembling all patch predictions yields the final segmentation. There are two typical methods for predicting the patch. On the ImageNet dataset, a PyTorch implementation of the Inception V3 architecture has been pretrained. Both the positive and negative areas are sampled beforehand. Extremely intense data augmentation was integrated and conducted during the training to improve the model's robustness (for example, flip and HSV colour argumentation). All patches were shrunk to the typical input size of Inception V3 of $299 \times 299$ pixels before being sent into the training procedure. The proposed model was implemented on a Dell T630 server equipped with 128GB of memory and 4 Titan X GPUs, and was run for 20 iterations. The Adam optimizer was used with a learning rate of $3 \times 10^{-4}$, and the data was divided into 92 batches. To preserve tissue architectural information and reduce computational cost, a patch size of $768 \times 768$ pixels was selected. The findings were comparable when we used a patch size of 384 × 384 pixels, but even with the patch size the computing cost was four times higher than the usual one.

In this paper [7], a deep learning approach for the segmentation and detection of colorectal cancer from digitized H&E-stained histology slides is presented. The results of this study show that, when compared to pathologist-based diagnosis using H&E-stained slides extracted from clinical samples, the neural network approach has a median accuracy of 99.9% for normal slides and 94.8% for cancer slides. Digital slides pertaining to 307 cases of colorectal cancer have been acquired. 222 slides had varied percentages of colorectal cancer, while 85 slides came from normal colon tissue. The entire slide is broken up into several patches, each of which is called separately. Assembling all patch predictions yields the final segmentation. There are two typical methods for predicting the patch. On the ImageNet dataset, a PyTorch implementation of the Inception V3 architecture has been pretrained. Both the positive and negative areas are sampled beforehand. Heavy data augmentation was integrated and conducted on the spot during training to improve the model's robustness (for example, flip and HSV colour argumentation). All patches were shrunk to the typical input size of Inception V3 of 299 × 299 pixels before being sent into the training procedure. The model ran for 20 iterations on a Dell T630 server with 128GB of memory and 4 Titan X GPUs. With Adam optimizer, the learning rate was set at $3 \times 10-4$. There were 92 batches in total. In order to maintain tissue architectural information and cut down on computing costs, we decided on a patch size of 768 × 768 pixels. The findings were comparable when we used a patch size of 384 × 384 pixels, but the computing cost was four times higher.

This study investigates the effect of re-sampling methods for unbalanced datasets on a PET radiomics-based prognostication model [8] for head and neck cancer patients. The model was applied to two patient cohorts, comprising 166 patients with newly diagnosed nasopharyngeal cancer (NPC) and 182 head and neck cancer patients from an open database. Overall survival (OS) and disease progression-free survival (DPFS) were analysed in relation to conventional PET parameters and robust radiomics characteristics (DFS) using a combination of 10 re-sampling techniques (oversampling, under sampling, and hybrid sampling) with 4 machine learning classifiers. The diagnostic performance in hold-out test sets was evaluated, and statistical differences were analysed using post hoc Nemenyi analysis with Monte Carlo cross-validations. The results showed that oversampling strategies such as ADASYN and SMOTE can improve prediction accuracy in terms of G-mean and F-measures for minority classes without significantly reducing F-measures for majority classes. The PET radiomics-based prediction model of OS for the NPC cohort was found to have an AUC of 0.82 and a G-mean of 0.77, and similar results were obtained when the model was tested on an external dataset. Additionally, oversampling approaches increased the prediction performance.

This study [9] intends to construct a deep convolutional neural network called CYENET for automatic cervical cancer diagnosis using colposcopy images. In this study, cervical lesions are identified from colposcopy pictures using a two-deep learning model. The CYENET architecture was created from scratch, and the transfer learning VGG 19 was adjusted for the suggested methodology. The input data size for the typical neural network architecture ranges from 1 by 1 to 5 by 5, and it uses a single type of CNN filter. The harmonic mean of the accuracy and reminder is used to compute the model's F1 score. The CYENET architecture was created from scratch, and the transfer learning VGG 19 was adjusted for the suggested methodology. High sensitivity, specificity, and kappa scores of 92.4%, 96.2%, and 88% were displayed by the suggested CYENET.

In order to categorize breast cancer cells, the suggested model [10] contrasts two machine learning algorithms—Naiye Bayes and K Nearest Neighbor—and uses the cross-validation technique to confirm each algorithm's accuracy. In cross-validation, the training and testing sets are randomly partitioned into 60% training sets and 40% testing sets, and they go through multiple crossovers rounds so that each instance is tested against. The simplest type is K-fold cross validation, which uses one of the K partitions as a validation set. construct accurate and trustworthy classifiers. Even if NB has an excellent accuracy rating of 96.19% and the noticed values of the KNN acquired a higher efficiency of 97.51% after a precise comparison between our algorithms, the KNN will take longer to run if the dataset is larger.

This study [11] applied three machine-learning algorithms – Support Vector Machine (SVM), K-nearest Neighbors (k-NN), and Decision Tree (DT) - to determine which classifier performs best in the categorization of breast cancer. The study used a duo-phase-SVM to evaluate the Wisconsin Breast Cancer Diagnosis (WBCD) dataset and achieved a classification model accuracy of 99.10% by combining a duo-phase clustering technique with a probabilistic SVM. When applied to the Wisconsin breast cancer cases dataset, the three algorithms - SVM, k-NN, and DT - were compared in terms of precision and training time. The results showed that quadratic SVM achieved an accuracy of 98.1% and outperformed the other classifiers.

III. PROPOSED SYSTEM

The system consists of three sections, namely, acquiring dataset, pre-processing, and AI Model.

The overview of our system starts with acquiring the dataset. We are collecting unprocessed images to simulate the real-life scenario. Following this step, noise and outliers are removed. The region of interest is further extracted from the resultant image. Pyradimics toolbox will further be used to look for features in the dataset. The ranked features are then input into the AI model to make a prediction.

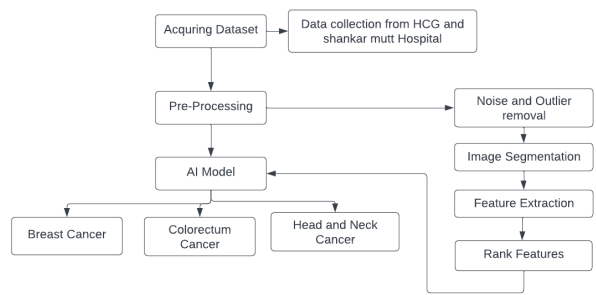

*A. Acquiring Dataset*

The dataset for the project will be acquired from different cancer hospitals. Mainly the dataset required for the project are requested from the shankar mutt cancer hospital in which the patient's data is hidden and only the raw microscopic images are given. Based on dataset and the type of cancer cell provided by the hospital we will be deciding on the type of cancer cell we will be classifying. We will mainly be looking for 3 types of cancer cells namely breast, colorectum, head and neck cancer cells. In case we were not provided with the dataset we will be looking either into existing datasets or request other hospitals for the same.

*B. Pre-Processing*

After the collection of datasets from hospital or different sources we will be applying different pre-processing algorithms to the images to extract features
in order to implement certain algorithms to classify the cancer cells as malignant or benign. The goal of preprocessing is to improve the quality of the image so that it can be more effectively analyzed. Preprocessing involves suppressing unwanted distortions and enhancing certain features that are relevant to the particular application at hand. The specific features to be enhanced may vary depending on the application. Preprocessing techniques can be grouped into four categories based on the size of the pixel neighborhood used to calculate a new pixel's brightness: (1) pixel brightness transformations, (2) geometric transformations, (3) techniques that use a local neighborhood of the processed pixel, and (4) image restoration techniques that require knowledge of the entire image. Some common preprocessing techniques include noise and outlier removal, image segmentation, feature extraction, and ranking of features. Preprocessing is applied in order to increase overall accuracy and reduce the error rate of the algorithm. For medical image processing, simple classification algorithms may not be sufficient, and it may be necessary to use pre-trained models and other algorithms to train and test the images.

*C. AI Model*

The proposed model will be the final results of comparison between multiple algorithms. The most used algorithms for medical images are k-Nearest Neighbor(k-NN), support vector machine, random forest and gradient boosting(XGBoost) machine. Every algorithm used has its own advantages and disadvantages in terms of accuracy, recall and F1 scores. This study employs five machine learning (ML) classifiers - support vector machine (SVM), k-nearest neighbor (kNN), random forest (RF), XGBoost, and LightGBM - to classify features extracted from breast ultrasound images as benign or malignant lesions. The ML models are optimized using the tree-structured Parzen estimator, and the dataset is divided using 10-fold cross-validation. The study includes the development of an algorithm for outlier detection, feature extraction using the pyradiomics toolbox, and a discussion and summary of ML and hyperparameter optimization. The most effective classifier for clinical application is identified and recommended for classifying bacteria cells as benign or malignant.

The above flowchart describes about the proposed model which follows the steps of taking the input features extracted from the image in the pre-processing stage. After the features are extracted the valid and primary features are visualized and selected. Then the model is implemented and trained on the pre-processed dataset are then compared with results based on all features and results based on selected features. The model selected would be expected to achieve high accuracy, high sensitivity, low error rate, good f1 score and it should be tested for different types of microscopic images if its malignant or benign.

## IV. CONCLUSION

In conclusion, the use of deep learning for cancer cell classification has gained significant attention in recent years as a promising approach for improving the accuracy and efficiency of cancer diagnosis and treatment. This survey paper reviewed the current state of the art in the use of deep learning for cancer cell classification, including a discussion of various deep learning architectures and techniques that have been applied in this context.
Overall, the findings from the studies reviewed in this paper suggest that deep learning can achieve high accuracy rates for cancer cell classification, and can be used to differentiate between different types of cancer cells. However, there are still challenges to be addressed in the use of deep learning for cancer cell classification, including the need for high-quality annotated data and the optimization of algorithms for clinical applications.
Despite these challenges, the use of deep learning for cancer cell classification holds great potential for advancing the field of cancer research and improving patient outcomes. Further research in this area is needed to fully realize the potential of deep learning for cancer cell classification, and to address the challenges that remain.